# Color: A Crucial Factor for Aesthetic Quality Assessment in a Subjective Dataset of Paintings

Seyed Ali Amirshahi,[1,2] Gregor Uwe Hayn-Leichsenring,[2] Joachim Denzler,[1] Christoph Redies[2]

[1] Chair of Digital Image Processing, Department of Computer Science, Friedrich Schiller University, Jena, Germany

[2] Experimental Aesthetics Group, Institute of Anatomy I, University of Jena School of Medicine, Jena University Hospital, Jena, Germany

**ABSTRACT**

Computational aesthetics is an emerging field of research which has attracted different research groups in the last few years. In this field, one of the main approaches to evaluate the aesthetic quality of paintings and photographs is a feature-based approach. Among the different features proposed to reach this goal, color plays an import role. In this paper, we introduce a novel dataset that consists of paintings of Western provenance from 36 well-known painters from the 15th to the 20th century. As a first step and to assess this dataset, using a classifier, we investigate the correlation between the subjective scores and two widely used features that are related to color perception and in different aesthetic quality assessment approaches. Results show a classification rate of up to 73% between the color features and the subjective scores.

## 1. INTRODUCTION

For centuries, philosophers and psychologists along with art critics have tried to find the bases of aesthetic judgment. In recent years, a new field of research named computational aesthetics has emerged (Amirshahi et al. 2012, Datta et al. 2006, Li and Chen 2009, Redies et al. 2012, Yanulevskaya et al. 2012). In this field, different research groups have tried to use computational techniques to identify image properties, which can be related to the aesthetic perception of a painting or a photograph. In this paper, we address two issues that are related to this field of research, i.e. subjective databases and color.

First, the availability of different photo-sharing websites allows researchers to have access to a huge number of high-quality photographs along with their subjectively rated aesthetic scores (Datta et al. 2006). Due to different copyright restrictions, no subjective database of aesthetic paintings is publically available. This prevents different metrics from being compared to one another. To the best of our knowledge, the only subjective dataset of aesthetic paintings which is intended for public use is the MART dataset (Yanulevskaya et al. 2012), which is limited to abstract paintings and, till now, has not been released to the public. In the present paper, we introduce a new subjective dataset of aesthetic paintings, which can be downloaded from the webpage dedicated to this dataset[1]. The paintings were originally included in the Yorck Project[2] or the Google Art Project[3] and were made available to the general public through the Wikimedia Commons website. In the present work, the paintings from the dataset were evaluated by non-expert observers to gather subjective scores on how beautiful they are.

---

[1] http://www.inf-cv.uni-jena.de/en/subjective_aesthetic_dataset

[2] http://commons.wikimedia.org/wiki/Category:PD-Art_(Yorck_Project)

[3] http://www.googleartproject.com

Second, it is a well-known fact that color plays an important role in an observer's aesthetic judgment. Different research has been carried out with regard to the role of color in aesthetic perception, based on general perceptual mechanisms as well as aesthetic quality assessment of paintings and photographs (Palmer et al. 2013). To study the relation between simple color features and subjective preferences for paintings, we focus on two color related features, color quantization and mean value for each color channel in the Lab color space. These measures have been previously associated with the aesthetic quality of paintings and photographs (Datta et al. 2006, Li and Chen 2009, Yanulevskaya et al. 2012).

In the next section, we will introduce the dataset. Section 3 is dedicated to evaluating the correlation between the color features and subjective scores. Finally, Section 4 gives a conclusion on the paper.

## 2. DESCRIPTION OF THE SUBJECTIVE DATASET

### 2.1 Collecting the Images

In the last few years, the Wikimedia Commons platform has provided users with high-quality digital reproductions of paintings, for which copyright has expired. From this database, we downloaded a dataset of 281 high-quality color paintings from 36 different artists. Careful attention was taken so that the downloaded images covered a wide variety of subject matters such as abstract, nearly abstract, landscapes, scenes with person(s), still life, flowers or vegetation, animals, seascape, port or coast, sky, portrait (one person), portrait (many person), nudes, urban scene, building, and interior scenes. The paintings are presumed to be of high aesthetic quality because they are works of well-known painters famous for their highly artistic skills. The webpage dedicated to this dataset[1] provides more information regarding the subject matters and art styles in this dataset as well as results of applying other aesthetic measures to this dataset.

### 2.1 Subjective Test

To collect the subjective scores, 49 participants (age: 19 to 44 years, M = 22.7 years, 13 males) were recruited. Most of the subjects were students, in particular of medicine or natural sciences. All of the participants declared having normal or corrected-to-normal visual acuity and gave their written informed consent after getting an explanation of the procedures. The study follows the ethical guidelines of the Declaration of Helsinki and was passed by the ethics committee of Jena University Hospital.

The downloaded pictures were resized to 2048 pixels on the longest side and no image enhancement algorithms were applied. In the experiment, images were shown separately and in a random way on a calibrated black screen (EIZO Color Edge CG241W). The size of the images on the screen was 165×165 mm. A chin rest assured a uniform viewing distance of 90 cm. The experiment was performed using Matlab R2008A. Prior to showing each picture, a question mark was presented (500 ms), followed by the image itself (600 ms) and a black screen (1900 ms). In that last phase, the participants were asked to respond by pressing one of four labelled keys. We asked the participants to rate the paintings on a scale from 1 (not beautiful) to 4 (very beautiful), according to how pleasant the picture appeared to them. The final score for each image was then calculated by averaging the scores collected from all observers. Figure 1 represents the 6 paintings with the highest and lowest average scores, respectively, along with a graph of the distribution of the ratings.

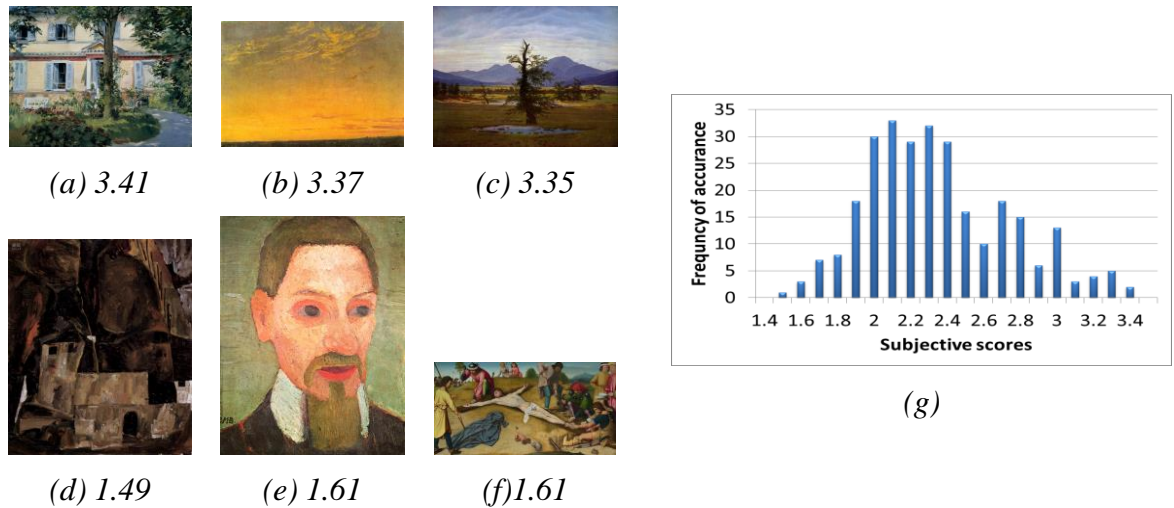

*Figure 1: Images with the highest and lowest subjective scores along with the distribution of the rating scores. a) Manet, 1882, b) Friedrich, 1824, c) Friedrich, 1822, d) Schiele, 1911, e) Modersohn-Becker, 1906, f) David, 1481, g) Distribution of the rating scores.*

## 3. RESULTS AND DISCUSSION

On average, observers tended to attribute lower aesthetic scores to paintings with brownish or dark colors while paintings with bluish and green colors received higher aesthetic scores. This finding is in agreement with the general color preferences described by Palmer and Schloss (2010). They showed that brownish and olive colors are more strongly disliked by Western adults, whereas blue and red colors are preferred by this group of observers.

To test whether simple color features can predict the aesthetic quality of a painting, two color features were extracted from the paintings, quantization of the color space into different colors and the overall mean values in different color channels in the Lab color space. These measures have previously been used in the work by Datta et al. (2006), Li and Chen (2009), and Yanulevskaya et al. (2012). In the present study, we quantized the Lab color space to 125 different colors (5 different levels for each color channel). This number was selected after testing different possible options. The images were then divided into two groups based on the subjective scores (paintings with high scores and low scores, respectively). To find the correlation between the subjective scores and the two mentioned color features, we trained an SVM classifier using 75% of the data and tested in on the remaining 25%. The two color features together resulted in a classification rate of 72.9% on average between five different cross-correlation image sets. Classification rates were 66.1% for the color quantization approach and 70.6% for the mean color feature alone.

## 4. CONCLUSION AND FUTURE WORKS

In this paper, we introduce a new subjective dataset of aesthetic paintings. As a first evaluation of this dataset, we find a high correlation between two previously used simple color features (color quantization and mean color value) and subjective beauty ratings of the paintings in the dataset. Results of the subjective score follow the pattern previously described by Palmer and Schloss (2010) with regard to liked and disliked colors. This result shows that, among different features used to evaluate the aesthetic quality of a painting or photograph, color plays a crucial role.

In future work, we will increase the number of images in our subjective dataset. Also, we will carry out a more in-depth analysis of the role of color in the aesthetic quality assessment of paintings by extracting other color features, for example features related to color harmony. Finally, we will combine multiple image features to extend the method of automatic aesthetic quality assessment of paintings (Li and Chen, 2009).